%% file: 0_Main.tex
\documentclass{article} % For LaTeX2e
\usepackage{iclr2021_conference,times}
\usepackage{makecell}
\usepackage{amsmath}
\usepackage{graphicx}
\usepackage{epstopdf} 
\usepackage{multirow}
\usepackage{times}
\usepackage{latexsym}
\usepackage{amssymb}
\usepackage{subfigure}
\usepackage{graphicx}
\usepackage[T1]{fontenc}
\usepackage{hyperref} 
\usepackage{bm}
\usepackage[utf8]{inputenc}
\usepackage{hyperref}
\usepackage{booktabs} 
\usepackage{colortbl}
\usepackage{microtype}
\usepackage[lined,boxed,commentsnumbered]{algorithm2e}
\usepackage{inconsolata}
\input{math_commands.tex}

\usepackage{hyperref}
\usepackage{url}
\iclrfinalcopy

\title{\includegraphics[height=0.7cm]{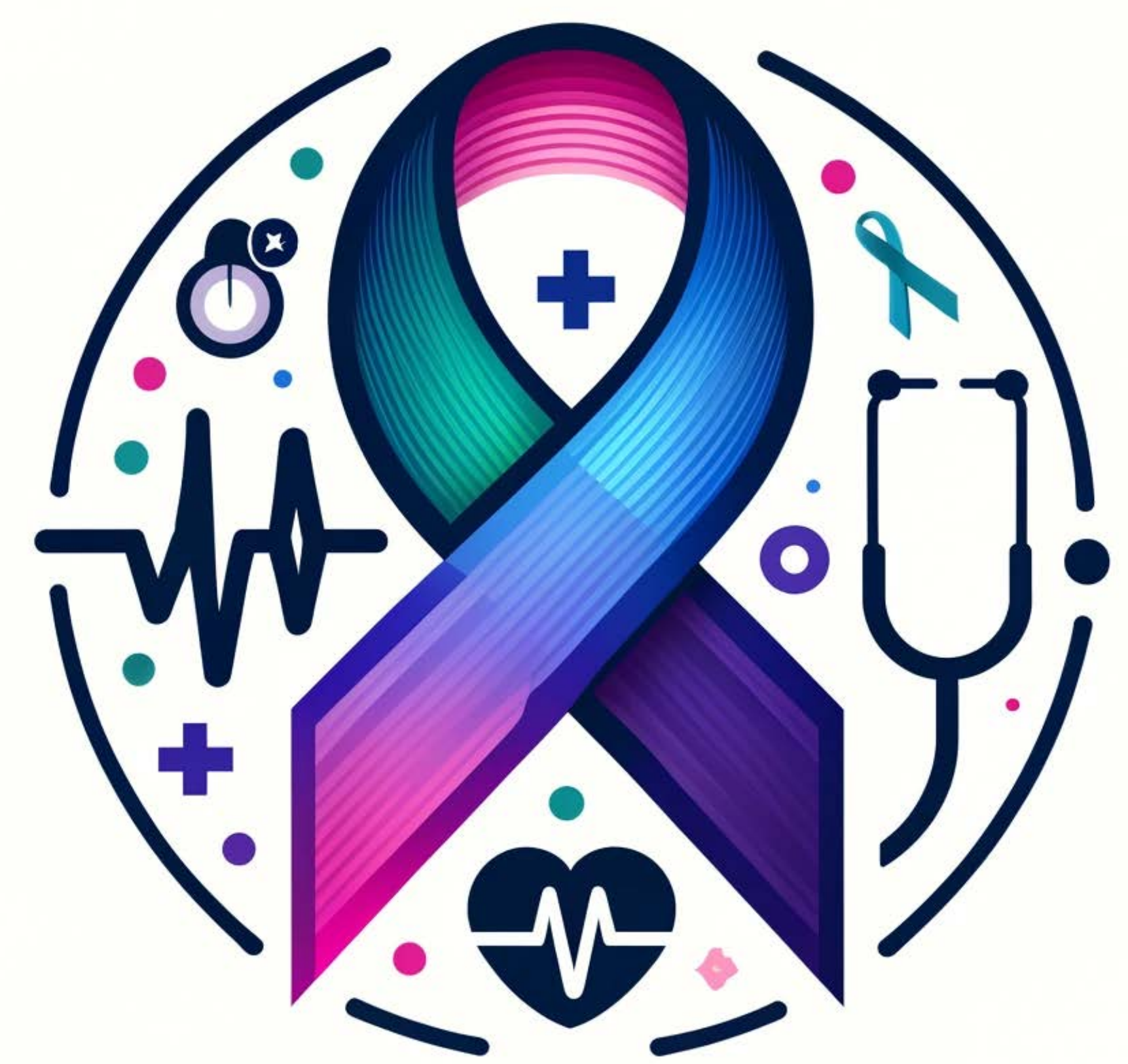}CancerLLM: A  Large Language Model in Cancer Domain}

\author{Mingchen Li\textsuperscript{\normalfont 1}, Zaifu Zhan\textsuperscript{\normalfont 2}, Jiatan Huang\textsuperscript{\normalfont 1}, Jeremy Yeung\textsuperscript{\normalfont 1}, Kai Ding\textsuperscript{\normalfont 3}, Anne Blaes\textsuperscript{\normalfont4} \\ \textbf{Steven Johnson}\textsuperscript{\normalfont5}, \textbf{Hongfang Liu}\textsuperscript{\normalfont6}, \textbf{Hua Xu}\textsuperscript{\normalfont7},\textbf{ Rui Zhang}\textsuperscript{\normalfont1} \\
\textsuperscript{1}Division of Computational Health Sciences, University of Minnesota Twin Cities\\ 
\textsuperscript{2}Department of Electrical and Computer Engineering, University of Minnesota Twin Cities\\
\textsuperscript{3}Department of Radiation Oncology and Molecular Radiation Sciences, Johns Hopkins
University\\
\textsuperscript{4}Division of Hematology, Oncology and Transplantation, University of Minnesota Twin Cities\\ 
\textsuperscript{5}Institute for Health Informatics, University of Minnesota Twin Cities\\ 
\textsuperscript{6}McWilliams School of Biomedical Informatics, UTHealth Houston\\ 
\textsuperscript{7}Department of Biomedical Informatics and Data Science, Yale School of Medicine\\ 
\textsuperscript{1}\{li003378, ruizhang\}@umn.edu\\
        }

\begin{document}

\maketitle
\input{1_abs}

\input{2_introduction}

\input{3_related_work}
\input{4_preliminaries}
\input{6_experiment}

\input{6_1_discussion}

\input{5_method}
\input{7_conclution}

% , go at the end of the paper.

\bibliography{iclr2021_conference}
\bibliographystyle{iclr2021_conference}

\appendix
% \section{Appendix}

\end{document}

%% file: math_commands.tex
\usepackage{amsmath,amsfonts,bm}

\def\eqref#1{equation~\ref{#1}}
\def\1{\bm{1}}

\DeclareMathAlphabet{\mathsfit}{\encodingdefault}{\sfdefault}{m}{sl}
\SetMathAlphabet{\mathsfit}{bold}{\encodingdefault}{\sfdefault}{bx}{n}

%% file: 1_abs.tex
\begin{abstract}
Medical Large  Language Models (LLMs) have demonstrated impressive performance on a wide variety of medical NLP tasks; 
however, there  still lacks a LLM specifically designed for phenotyping identification and diagnosis in cancer domain. Moreover, these LLMs typically have several billions of parameters, making them computationally expensive for healthcare systems. 
Thus, in this study, we propose \textbf{CancerLLM},
a model with 7 billion parameters and a Mistral-style architecture, pre-trained on nearly 2.7M clinical notes and over 515K pathology reports covering 17 cancer types, followed by fine-tuning on two cancer-relevant tasks, including cancer phenotypes extraction and cancer diagnosis generation. Our evaluation demonstrated that the CancerLLM achieves state-of-the-art results with F1 score of 91.78\% on phenotyping extraction and 86.81\% on disganois generation. It outperformed existing LLMs, with an average F1 score improvement of 
\textit{9.23}\%. Additionally, the CancerLLM demonstrated its efficiency on time and GPU usage, and robustness comparing with other LLMs.
%%%%%%%%%%
We demonstrated that CancerLLM can potentially provide an effective and robust solution to advance clinical research and practice in cancer domain.
% This illustrates that smaller, cancer-targeted models have the potential to serve as cost-effective and proficient frameworks for various cancer applications, potentially aiding healthcare professionals.
\end{abstract}

%% file: 2_introduction.tex
\section{Introduction}
% The remarkable success of the Large Language Model(LLMs) across diverse NLP tasks

Recently, large language models (LLMs) like GPT-4\footnote{https://openai.com/index/gpt-4/}, and LLama2~\citep{touvron2023llama}  have become the dominant technology in various natural language processing (NLP) tasks, these models develop impressive capabilities in different NLP tasks, including information extraction (e.g., \citep{li2023far,li2024biomedrag,li2024benchmarking,zhang-etal-2024-prompt-tuning-shot}), link prediction (e.g., \citep{li2024condensed,liu2024behaviornet,li2022hierarchical}), and question answering (e.g., \citep{zhuang2024toolqa,jiang2024enhancing,huang2024ritek}). While much attention has been given to these models' capabilities in the general domain, it's evident that specialist models have the potential to significantly help clinical and biomedical research~\cite{singhal2023large}. To tailor LLMs for the clinical domain, several specialized models have been developed, such as ClinicalCamel 70B~\citep{toma2023clinical} and Llama3-OpenBioLLM 70B\footnote{https://huggingface.co/aaditya/Llama3-OpenBioLLM-70B}. Improving domain-specific language models will accelerate breakthroughs in clinical and biomedical research, leading to enhanced patient care.

Despite showcasing impressive general capabilities, current medical LLMs face significant challenges when applied to cancer~\citep{yang2023large,ullah2024challenges,zhou2024large}. This is primarily due to a deficiency in cancer-specific knowledge within these models~\citep{perez2024guide,truhn2023large}. This lack of specialized understanding poses a barrier to the effective utilization of LLMs in assisting doctors with cancer diagnosis and treatment planning.
%%%%%%%%%%%%
Expanding cancer-specific knowledge within LLMs is imperative for surmounting these obstacles and harnessing their full potential in the cancer field~\citep{tsang2025foundation}. By integrating comprehensive information encompassing aspects like cancer diagnosis, subjective information, objective information, nursing review of systems (nursing ros) records, laboratory test results, patient self-descriptions, medical history, treatment modalities, assessments, and patient outcomes, LLMs can furnish healthcare professionals with more precise and personalized recommendations~\citep{alsaad2024multimodal}. This augmented capability holds the promise of revolutionizing cancer care and aiding medical professionals in enhancing diagnostic accuracy and formulating treatment plans. 
%%%%%%%%%
To the best of our knowledge, CancerBERT~\citep{zhou2022cancerbert} is the only model specifically designed for the cancer domain. While CancerBERT~\citep{zhou2022cancerbert} stands out as a specialized model for cancer-related tasks, its focus is primarily on breast cancer, leaving other cancer types relatively unaddressed. This limitation underscores the need for further development of specialized models that cover a broader spectrum of cancer types and applications. By expanding the scope of such models to encompass various cancer types and different applications, healthcare professionals can access more comprehensive and tailored resources to enhance patient care in the cancer domain.

Except for the lack of knowledge, to evaluate the capability of LLMs in the cancer domain, the lack of high-quality datasets grounded in real-world electronic health records (EHRs) remains a major challenge~\citep{zaied2015electronic,dar2022breast}. Most existing EHR datasets are proprietary and restricted due to privacy concerns, limiting the community's ability to systematically evaluate LLMs on clinically meaningful tasks. As such, constructing dedicated datasets tailored to cancer-specific applications is essential for enabling accurate, task-relevant assessment of model capabilities~\citep{varlamova2024machine,vorontsov2024foundation}. These datasets play a pivotal role in benchmarking LLM performance under realistic clinical conditions and guiding the development of domain-adapted models that can truly support oncology care.

Moreover, the deployment of large parameter LLMs presents a significant challenge for hospitals or medical institutions with limited computational resources. Models with a dozen billions of parameters, such as those 
models with 13 billion parameters, like PMC LLaMA 13B~\citep{wu2023pmc} and Medalpaca 13B~\citep{han2023medalpaca}, or those with 70 billion parameters, such as Llama3-OpenBioLLM-70B\footnote{https://huggingface.co/aaditya/Llama3-OpenBioLLM-70B} and ClinicalCamel-70B~\citep{toma2023clinical}, pose significant computational challenges for hospitals or medical institutions with limited resources, require substantial computing power and infrastructure to train and deploy effectively. The financial and technical constraints associated with running these models may pose barriers to their widespread adoption in cancer settings. As such, there is a pressing need to develop a smaller LLM for the cancer domain, reducing computational requirements to ensure that advanced LLMs can benefit a broader range of medical professionals and patients.

% Despite the rapid progress of instruction-tuned language models, high-quality datasets tailored for critical clinical tasks such as phenotype extraction and diagnosis generation remain limited.

% To address this gap, we develop two specialized instruction-tuning datasets focused on cancer phenotype extraction and diagnosis generation.

\begin{figure*}[t]
        \centering
        \includegraphics[width=1\columnwidth]{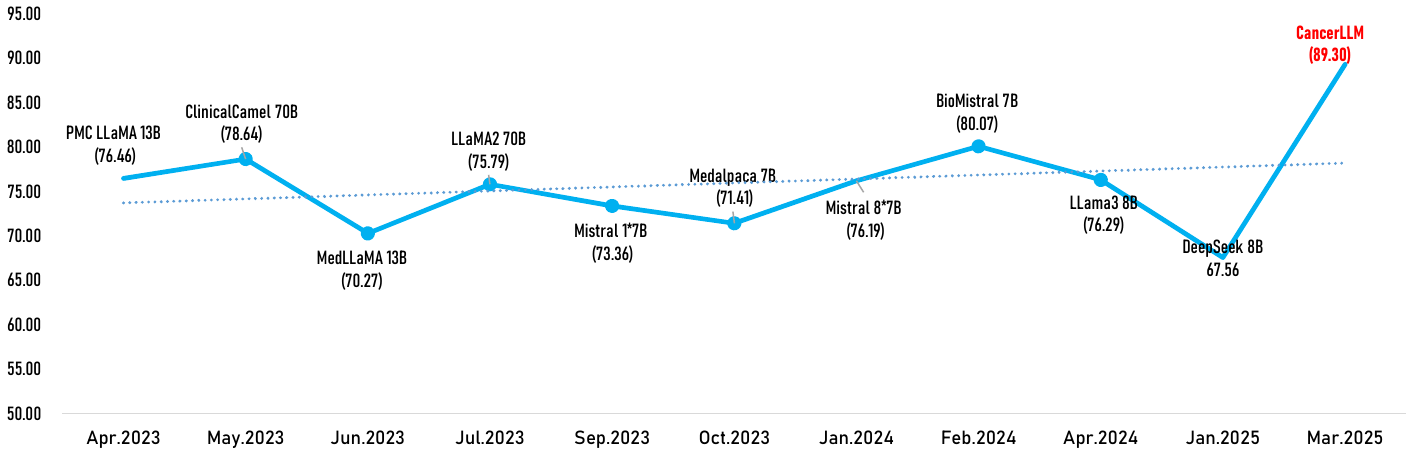}
	\caption{The evolution of medical LLM performance on cancer phenotype extraction, and diagnosis generation is measured using the average F1 score, which includes Exact Match, BLEU-2, and ROUGE-L. Our \textsc{CancerLLM} achieves the highest performance with an F1 score of 89.30\%. } 
	\label{con:evolution}
\end{figure*}

To bridge these gaps,
we first construct two datasets focused on cancer phenotype extraction and diagnosis generation. 
We then introduce CancerLLM, a state-of-the-art 7B-parameter language model tailored to enhance LLM proficiency in the cancer domain and support medical professionals in both phenotype extraction and diagnosis generation tasks.
Our model represents an important step towards democratizing advancements in the cancer domain of LLMs. More specifically, our research focuses on four steps: 1) pre-training the LLM in the style of Mistral. We pre-train the Mistral 7B  using 2,676,642 cancer clinical notes and 515,524 pathology reports from the University of Minnesota (UMN) Clinical Data Repository. 2) fine-tuning the LLM on a specialized dataset.
We proposed two datasets focusing on phenotype extraction and diagnosis generation to evaluate the information extraction and diagnosis ability of CancerLLM. Following this, we fine-tuned CancerLLM using instruction learning for the aforementioned tasks.  3) We evaluate the generation ability of CancerLLM against other medical LLMs using Exact Match~\citep{rajpurkar2016squad}, BLEU-2~\citep{papineni2002bleu}, and ROUGE-L~\citep{cohan2016revisiting} metrics. 4)  To further evaluate the model's robustness, we propose two robustness testbeds, which include counterfactual robustness and misspellings robustness in the special tasks. 5) To examine the effectiveness of different examples in instruction tuning, we proposed retrieval-based CanerLLM, and further evaluated the task performance of five retrieval-based models built upon the Cancer LLM.

As shown in Figure ~\ref{con:evolution}, our model not only outperforms existing 7B and 13B models by a significant margin but also delivers good results with those of 70B models.

The contributions of this work can be summarized as follows:
\begin{itemize}
     \item We proposed \textsc{CancerLLM}, a large language model which focuses on the cancer domain, as well as retrieval-based CancerLLM with five retrieval-based models. 
     \item  An original study with the introduction of a benchmark of phenotype extraction and diagnosis generation, facilitating
assessment against existing state-of-the-art open-source medical LLMs.
    \item  We conducted an  in-depth quantitative analysis of the model’s generation ability and robustness on proposed two testbeds on phenotype extraction and diagnosis generation.
\end{itemize}

%% file: 6_experiment.tex
\section{Results}
\label{experiments}
\subsection{Main Results}

\begin{table*}[ht]
	\centering
	\renewcommand\arraystretch{1.3}
\resizebox{0.9\textwidth}{!}{%
	\begin{tabular} {l|ccc|ccc|ccc|c}
		\toprule 
		\multicolumn{1}{c}  {}&\multicolumn{3}{c}  {Exact Match} &\multicolumn{3}{c}  {BLEU-2}&\multicolumn{3}{c}  {ROUGE-L}&\multicolumn{1}{c}  {F1}\\
		%\hline
		 Approach & Precision &  Recall & F1 & Precision &  Recall & F1 & Precision &  Recall & F1 &  Average \\
	      \midrule
        PMC LLaMA 7B&  47.04   &   47.04   &    47.04 &   56.41    & 56.41   &56.41    & 66.98 & 66.98  & 66.98 &  56.81\\
         Medalpaca 7B &   41.75   &  41.75   &   41.75  &   50.16  & 50.16   & 50.16    &62.07  &62.07  &62.07  &51.33\\
          LLAMA-2 7B &  33.18    &     33.18   &   33.18   & 42.80   &  42.80 &   42.80   & 55.09 & 55.09& 55.09   &43.96 \\
   Mistral 1*7B& 45.40    & 45.40 &   45.40 &  54.29    & 54.29   &   54.29   &  65.52  &  65.52&  65.52  &  55.07\\
     Mixtral 8*7B& 51.32     &51.32         &   51.32    & 59.35    & 59.35  & 59.35    & 69.70  & 69.70     & 69.70  &  60.12\\
      Bio-Mistral 7B & 62.26   & 62.26   &  62.26  &  68.40   &  68.40  &  68.40     &76.02  &76.02   & 76.02 &68.89\\
 LLama3 8B&  51.60  &  51.60   &    51.60  &  60.13    &  60.13  &  60.13    &  69.97   & 69.97    &  69.97  & 60.57 \\
 
 Qwen-7B&     76.66   &   76.66    &   76.66   & 80.59   &  80.59     &  80.59    &   85.46  &  85.46  & 85.46  & 80.90 \\
  Deepseek 8B&  34.82   &    34.82 &   34.82    &  44.98    & 44.98     &    44.98   &  56.97   &  56.97     &   56.97   & 45.59  \\
                \hline
                MedLLaMA 13B &   39.02   &39.02     & 39.02     &48.98       &  48.98 &  48.98    &60.63   &60.63 &60.63    &49.54 \\
                PMC LLaMA 13B  &    54.97   & 54.97     &  54.97       &62.56      &  62.56    & 62.56       & 71.06   &  71.06   &  71.06   &62.86\\
                Medalpaca 13B  &    40.66    & 40.66    &   40.66    & 50.58      & 50.58   &   50.58        &62.40   & 62.40   &  62.40   & 51.21 \\
                 LLaMA2 13B & 45.76   &   45.76     &  45.76    &  54.70     &54.70   &  54.70   &   65.80  &65.80   & 65.80  & 55.42\\
                 \hline
                  LLaMA2 70B &50.23     &  50.23        &  50.23    & 59.63     & 59.63  &   59.63  & 69.25   &  69.25  &  69.25 & 59.70\\
                % MEDITRON-70B~\cite{chen2023meditron} &      &        &      &     &   &     & \\
                Llama3-OpenBioLLM-70B&   54.42     &   54.42       & 54.42      &   62.36   & 62.36 &   62.36 & 71.65  &  71.65 &  71.65 &62.81\\
                 ClinicalCamel-70B& 54.60    &   54.60       &   54.60   &  63.34   & 63.34 &   63.34   & 72.73   & 72.73&72.73 & 63.55\\
                \hline
                % \textsc{CancerLLM$^1$}(Ours) &  79.85    &  79.85      & 79.85   &  83.66    &83.66  &     83.66 &  88.51  & 88.51    &  88.51   &  84.00 \\
                 \textsc{CancerLLM 7B}(Ours)&   \textbf{83.50}    &   \textbf{83.50 }      &   \textbf{83.50}    & \textbf{86.60}      &\textbf{86.60}   &  \textbf{86.60}    &   \textbf{90.34}    &\textbf{90.34}   &\textbf{90.34}   &   \textbf{86.81}\\
                 %     Ours-test2 (pre-training biomixtural, fine-tuning: mixtrual ) &  79.58     &    79.58   &   79.58   &   83.67    &   83.67     &   83.67       &  88.40   &  88.40    &  88.40    &  83.88   \\
                 % Ours-test3 (pre-training biomixtural, fine-tuning: biomixtrual ) &  79.49       &    79.49   & 79.49     &     83.74   &  83.74    &     83.74   & 88.67  & 88.67    &88.67     &  83.97   \\
           % Med-Alpaca &     &     &   \\
           \bottomrule
	\end{tabular}
  }
\vspace{+2mm}
% , In CancerLLM$^1$, the model in the fine-tuning process using Biomixtrial 7B, while In CancerLLM$^2$, the model in the fine-tuning process using Mixtrial 7B
\caption{Results of   Cancer Diagnosis Generation} 
\label{con:diagnosis_generation}
\vspace{-1mm}
\end{table*}
\subsubsection{Cancer Diagnosis Generation}
To assess the scalability of our model, in this part, we evaluate the performance of \textsc{CancerLLM} in diagnosis generation task. 
Table~\ref{con:diagnosis_generation} presents the experiment results of various medical LLMs. We have the following observations: (1) our \textsc{CancerLLM} significantly outperforms all the strong baselines and its variants across all evaluation metrics. (2) We observed that \textsc{CancerLLM } improve the original   \textsc{Mistral 1*7B}, and \textsc{Bio-Mistral 7B}   by  31.7\%, and 17.92\%  respectively, in term of average F1 across Exact Match, BLEU--2 and ROUGE-L. (3) 
Our model not only achieves the best performance among 7B models, but also outperforms medical LLMs with 13B and 70B parameters, demonstrating that a lightweight, domain-specific model can effectively rival much larger general-purpose models.

\begin{table*}[ht]
	\centering
	\renewcommand\arraystretch{1.3}
\resizebox{0.9\textwidth}{!}{%
	\begin{tabular} {l|ccc|ccc|ccc|c}
		\toprule 
		\multicolumn{1}{c}  {}&\multicolumn{3}{c}  {Exact Match} &\multicolumn{3}{c}  {BLEU-2}&\multicolumn{3}{c}  {ROUGE-L}&  F1\\
		%\hline
		 Approach & Precision &  Recall & F1 & Precision &  Recall & F1 & Precision &  Recall & F1 & Average\\
	      \midrule
        % Mamba 2.7 B~\cite{gu2023mamba} &     &     &    &     &    &     &       \\
        PMC LLaMA 7B  &   88.61    &     88.61    &  88.61      & 90.36      &  90.36   &     90.36   & 93.26   &  93.26  &  93.26   &  90.74\\
         Medalpaca 7B  &   89.28     &  89.28      &  89.28     &   91.27    &  91.27    &  91.27       &93.89   & 93.89   & 93.89    &91.48  \\
          LLAMA-2 7B  &  89.18     &   89.18        &  89.18        &  90.83       &   90.83     &   90.83       &  93.53  &93.53      & 93.53    & 91.18 \\
   Mistral 1*7B  &    89.47     &     89.47      &   89.47       &   91.30   & 91.30     & 91.30        & 94.19    &  94.19  &   94.19 & 91.65\\
     Mixtral 8*7B  &   90.23      &     90.23    &     90.23   &    92.05    &92.05      &   92.05      & 94.50     & 94.50     &94.50       &  92.26 \\
      Bio-Mistral 7B &   88.90     &    88.90     &     88.90  & 91.05      &91.05      &  91.05        &   93.79   &93.79    &  93.79  &  91.24 \\
 LLama3 8B  &    89.94    &    89.94    &  89.94     &  91.75     &   91.75      &   91.75         & 94.34   & 94.34   &   94.34 & 92.01\\
 
   Qwen-7B&     88.90    &88.90       &   88.90   &   90.67  &  90.67       &    90.67    &   93.57   &  93.57    &    93.57  & 91.05\\
    Deepseek 8B&   87.19   &87.19    &  87.19     &  89.12    &  89.12     &   89.12     &92.24     &92.24     & 92.24   &  89.52 \\
                \hline
                MedLLaMA 13B  &    88.80    &    88.80    &  88.80     &   90.87    &  90.87     &  90.87      & 93.31     &  93.31  &  93.31    &  90.99\\
                PMC LLaMA 13B &   87.95    &  87.95           &   87.95         &  89.64     &   89.64    &  89.64        &  92.58   &  92.58 &92.58   &  90.06\\
                Medalpaca 13B  &  88.61      &   88.61        &    88.61     & 90.37       &    90.37 &  90.37         &92.94     &  92.94 
 &  92.94 & 90.64 \\
                 LLaMA2 13B &    89.85    &   89.85     &    89.85    &  91.54      &  91.54      & 91.54         &  94.21   &   94.21    &   94.21  & 91.86\\
                 \hline
                  LLaMA2 70B  &    90.04    &    90.04    &   90.04    & 91.62     &91.62     &   91.62      &93.98     &93.98     & 93.98    &   91.88\\
                % MEDITRON-70B~\cite{chen2023mtestingeditron} &      &        &      &     &   &     & \\
                    Llama3-OpenBioLLM-70B&  88.33       &   88.33      &  88.33    &   90.02    &           90.02  & 90.02  &   93.15  &  93.15  &   93.15     & 90.50 \\
                ClinicalCamel-70B  &  \textbf{92.02}      &  \textbf{92.02}        &    \textbf{92.02 }   &   \textbf{93.62}    &  \textbf{ 93.62}   &    \textbf{93.62}     &  \textbf{95.52}   &  \textbf{95.52 } &   \textbf{95.52}  &  \textbf{93.72}  \\
                \hline
                  CancerLLM 7B(Ours)&    \underline{89.37}    &   \underline{89.37 }      &   \underline{89.37}      &    \underline{91.98}     & \underline{91.98}    &   \underline{91.98}     &   \underline{93.98}  &  \underline{93.98 }  &    \underline{93.98}  &  \underline{91.78 } \\
                 % CancerLLM$^1$(Ours)&    \underline{89.37}    &   \underline{89.37 }      &   \underline{89.37}      &    \underline{91.98}     & \underline{91.98}    &   \underline{91.98}     &   \underline{93.98}  &  \underline{93.98 }  &    \underline{93.98}  &  \underline{91.78 } \\
                    % CancerLLM$^2$(Ours) &  88.90    &   88.90       &  88.90     &   91.20      &   91.20  &   91.20    &  93.81  & 93.81   &    93.81  & 91.30   \\
                 
                 %  Ours-test2 (pre-training biomixtural, fine-tuning: mixtrual ) & 84.06       &  84.06      &  84.06     & 86.35       &  86.35      &    86.35      &   88.76 &    88.76 &   88.76  &86.39     \\
                 % Ours-test3 (pre-training biomixtural, fine-tuning: biomixtrual ) & 88.90      & 88.90      & 88.90     & 91.20       &    91.20   &      91.20   &  93.81   & 93.81     & 93.81     & 91.30    \\    
           % Med-Alpaca &     &     &   \\
           \bottomrule
	\end{tabular}
  }
\vspace{+2mm}
% ,  CancerLLM$^1$ refers to the model in the fine-tuning process using Biomixtrial 7B, while CancerLLM$^2$ refers to the model in the fine-tuning process using Mixtrial 7B. Bold font indicates the best model performance, while underline font represents our model's performance
\caption{Results of  Cancer Phenotype Extraction }
\label{con:Model_performance_phenotype}
\vspace{-1mm}
\end{table*}

\subsubsection{Cancer Phenotype Extraction}

Table~\ref{con:Model_performance_phenotype} presents the experiment results of various approaches on the treatment plan generation task. We have the following observations: 
(1) Our \textsc{CancerLLM} significantly outperforms all strong baselines with the same parameter numbers or number of LLMs across all evaluation metrics.
(2) Our \textsc{CancerLLM} performs comparably with LLMs that have larger parameters. For instance, LLama3 8B achieves 92.01\%, and LLama2 13B achieves 91.86\%.
(3) \textsc{CancerLLM}  could improve the performance of original   \textsc{Mistral 1*7B}, and \textsc{Bio-Mistral 7B}  in term of average F1 across Exact Match, BLEU--2 and ROUGE-L.  (4) We observe that ClinicalCamel-70B achieves the best phenotype extraction performance; however, it has a larger number of parameters, which affects both training and inference time, as well as memory consumption.
(5) Despite our \textsc{CancerLLM} not showing significant improvement across all LLMs with different parameters on the phenotype extraction task, as depicted in Figure~\ref{con:evolution}, our model attains the best performance across the three tasks examined in this paper.

\subsection{Results of Robustness Testbeds}

\subsubsection{Counterfactual Robustness}
% \begin{figure*}[t]
%         \centering
%         \includegraphics[width=0.5\columnwidth]{phenotype_extraction.pdf}
% 	\caption{ Average F1 results across different counterfactual rate of  \textsc{CancerLLM} and Mistral 1*7B.} 
% 	\label{con:conterfacutal_rate_results}
% \end{figure*}

\begin{table*}[ht]
	\centering
	\renewcommand\arraystretch{1.3}
\resizebox{0.9\textwidth}{!}{% s
	\begin{tabular} {l|l|ccc|ccc|ccc|c}
		\toprule 
		\multicolumn{1}{c}  {}&\multicolumn{1}{c}  {}&\multicolumn{3}{c}  {Exact Match} &\multicolumn{3}{c}  {BLEU-2}&\multicolumn{3}{c}  {ROUGE-L}&\multicolumn{1}{c}  {}\\\
		%\hline
		 Model &rate & Precision &  Recall & F1 & Precision &  Recall & F1 & Precision &  Recall & F1 & Average F1  \\
	      \midrule
           % \multirow{4}*{  Mistral 1*7B  }  &   20\%  &  89.85     & 89.85      &   89.85    &91.95     &  91.95   &   91.95  & 94.35   &94.35     & 94.35 \\
           \multirow{5}*{  Mistral 1*7B  }  &   20\%  &  88.01     & 88.01       &  88.01    &89.20    &  89.20   &   89.20 & 94.10  &94.10     & 94.10& \multirow{4}*{  72.53 } \\
         &    40\%   &    84.82   &  84.82     &   84.82    & 86.85     & 86.85    &   86.85  &  89.58   &  89.58    & 89.58 &  \\
         &   60\%   &    77.13     &  77.13      &   77.13    &  79.49  &  79.49  &    79.49&  82.66    &  82.66 &82.66 &   \\
           &   80\%    &  31.97     &   31.97     &    31.97    &   32.29   & 32.29   &32.29   & 34.24   &    34.24  & 34.24  &  \\
        % &       &       &        &       &      &    &    &    &       &  &  \\
           % & 0 &    89.47     &     89.47      &   89.47       &   91.30   & 91.30     & 91.30        & 94.19    &  94.19  &   94.19 & 91.65  \\
           \hline
       \multirow{4}*{ CancerLLM(Ours) }&   20\%  &  88.05     & 88.05     & 88.05     & 89.90    &  89.90   & 89.90    &  92.26   &   92.26   & 92.26 &\multirow{4}*{ \textbf{ 78.03} }  \\
         &    40\%   &    84.54   &    84.54    &  84.54      &  86.71    &   86.71   &   86.71   & 89.56   &    89.56  & 89.56 &   \\
         &   60\%   &     77.51  &    77.51     &  77.51       &  79.18    &    79.18   &   79.18    & 82.26   &   82.26  &82.26 &  \\
           &   80\%    &   54.55    &    54.55   &     54.55  & 55.39    &   55.39   &   55.39   &  56.53  &  56.53     &  56.53 &  \\ %CancerLLM$^1$(Ours) 
                   % &       &       &        &       &      &    &    &    &       &  & =\textbf{78.03} \\
           \bottomrule
	\end{tabular}
  }
\vspace{+2mm}
\caption{Counterfactual robustness performance on phenotype extraction. The rate refers to different counterfactual rate.}
\label{con:robustness_performance_test_bed1}
\vspace{-1mm}
\end{table*}
To validate the robustness of our model, we propose a counterfactual robustness testbed designed to simulate incorrect annotations. 
Table~\ref{con:robustness_performance_test_bed1}  presents the complete results under various counterfactual rates. We observed that: (1) Our CancerLLM still achieves the best performance when compared to Mistrial 1*7B. 
(2) As the rate increases, the model's performance deteriorates. (3) 
When the rate is set as 20\%, 40\%, and  60\%, our model exhibits similar performance to Mistral 1*7B. However, when the rate is set at 80\%, we observe that CancerLLM achieves higher F1 performance, indicating the robust counterfactual resilience of our model.

\subsubsection{Misspellings Robustness}
% \begin{figure*}[t]
%         \centering
%         \includegraphics[width=0.5\columnwidth]{diag_treatment.pdf}
% 	\caption{ Average F1 results across different misspelling rate of  \textsc{CancerLLM} and Bio-Mistral 7B} 
% 	\label{con:diag_treatment_ro_test_bed2}
% \end{figure*}
\begin{table*}[ht]
	\centering
	\renewcommand\arraystretch{1.3}
\resizebox{1\textwidth}{!}{%
	\begin{tabular} {l|l|l|ccc|ccc|ccc|c}
		\toprule 
	\multicolumn{1}{c}  {}&	\multicolumn{1}{c}  {}&\multicolumn{1}{c}  {}&\multicolumn{3}{c}  {Exact Match} &\multicolumn{3}{c}  {BLEU-2}&\multicolumn{3}{c}  {ROUGE-L}&\multicolumn{1}{c}  {}\\
		%\hline
		 &Model &rate & Precision &  Recall & F1 & Precision &  Recall & F1 & Precision &  Recall & F1 & Average F1\\
	      \midrule
         \multirow{8}*{ Diagnosis Generation } & \multirow{4}*{ Bio-Mistral 7B }  &   2\%   & 0.82       &  0.82       &    0.82     &  10.38    & 10.38     & 10.38   & 25.23    &  25.23     & 25.23&\multirow{4}*{  12.06}  \\
            &  &   4\%   &  0.64     &  0.64        &0.64          &  11.78     &   11.78    &    11.78 &  27.01   & 27.01      & 27.01 &\\
              &  &   6\%   & 0.00      &   0.00     &     0.00   &    10.23   & 10.23     &  10.23  &  25.25    &     25.25   & 25.25 & \\
               &  &   8\%   &  0.09      & 0.09         &   0.09       & 9.56      &  9.56     & 9.56    &  23.80   & 23.80      &  23.80 & \\
         %    & &   20\%  &  1.09     &    1.09    &    1.09    &  11.77    &   11.77  &   11.77  &  27.56  &  27.56    & 27.56 & \\
         %  & &    40\%   &  0.18     &   0.18    &  0.18  & 7.40    &7.40    &   7.40 &   21.48   &21.48& 21.48 &  \\
         % &  &   60\%   &    0.00  &   0.00     &    0.00    &   3.86   &    3.86 &    3.86 &  19.36   &   19.36  & 19.36  &\\
         %   &  &   80\%    &  0.00    & 0.00      &  0.00     &1.14      & 1.14   &    1.14& 15.87   & 15.87     &  15.87    & \\
              \cmidrule(rr){2-13}
       & \multirow{4}*{ \textsc{CancerLLM}(Ours) } &   2\%  & 0.09       &   0.09       &    0.09      &   9.88   & 9.88       &  9.88    &  24.67   &   24.67    &  24.67&\multirow{4}*{ 12.00} \\  %\textsc{CancerLLM$^2$}(Ours)
            &  &   4\%   &     0.00    &  0.00        &   0.00       &     10.29  &    10.29   &   10.29  &   25.37       &25.37 &25.37 & \\
              &  &   6\%   &    0.00    & 0.00        &   0.00      & 9.54      &   9.54   &   9.54 &  24.16    & 24.16       &24.16  & \\
               &  &   8\%   & 0.27      & 0.27        &     0.27    &   10.61      &  10.61    &     10.61  & 25.81      & 25.81  &25.81&\\

           \bottomrule
	\end{tabular}
  }
\vspace{+2mm}
\caption{Misspellings robustness performance on diagnosis generation. The rate refers to different misspelling rates in each instance.}
\label{con:diag_treatment_ro_test_bed2_whole_result}
\vspace{-1mm}
\end{table*}

Table~\ref{con:diag_treatment_ro_test_bed2_whole_result} presents the robustness evaluation of CancerLLM and Bio-Mistral 7B under varying levels of misspellings in the diagnosis generation task. The misspelling rates range from 2\% to 8\%, and the table reports performance using Exact Match, BLEU-2, and ROUGE-L metrics. Both models show a drop in performance as the misspelling rate increases, which is expected. Bio-Mistral 7B achieves slightly higher average F1 scores at lower error rates, especially at 2\% and 4\%. However, CancerLLM demonstrates more consistent performance across all levels, notably outperforming Bio-Mistral at the 8\% rate in terms of Exact Match (0.27\% vs. 0.09\%) and ROUGE-L (25.81\% vs. 23.80\%). The overall average F1 scores of both models are very close, with Bio-Mistral 7B at 12.06\% and CancerLLM at 12.00\%, indicating comparable robustness. These results suggest that CancerLLM maintains stable generation quality under noisy clinical inputs, validating its reliability in real-world settings

\subsection{Results of Generation Time and GPU Memory}

\begin{table}[ht]
	\centering
	\renewcommand\arraystretch{1.3}
	\scalebox{0.7}{
	\begin{tabular} {cccccccccc}
		\hline 
		 &  \multicolumn{3}{c} {Phenotype Extraction} &  \multicolumn{3}{c} {Diagnosis Generation }\\ 
		\hline
   LLMs& F1 & Time& Used  Memory  & F1 &  Time & Used Memory\\
   \hline	
      Bio-Mistral 7B &91.24   &1:06:55& 5,746 MB  & 68.89 &   1:07:45&5,802 MB \\
      Mistral 1*7B &91.65  & 57:05&5,598 MB &      55.07   &  1:07:49&5,680 MB \\
      Mixtral 8*7B & 92.26 & 2:01:27&25,086 MB &    60.12 &   2:16:14&25,166 MB \\
      PMC LLaMA 13B& 90.06 &1:08:59&8,208 MB &      62,86& 1:19:52&9,208 MB  \\
      LLaMA2 13B& 91.86& 1:08:43&8,204 MB &    55.42 &    1:24:17& 9,254 MB \\
       ClinicalCamel-70B&93.72   & 2:50:16&37,716 MB&       63.55&   3:05:37&37,67 MB \\
         \hline
         \textsc{CancerLLM}(Ours)&91.78   &1:14:12& 5,550 MB&     86.81 &1:26:33&5,768 MB  \\
		\hline
	\end{tabular}
 }
		\caption{Comparation of Generation Time (hours:minutes:seconds) and Used GPU Memory (Used Memory: Megabyte (MB)) of different LLMs, F1 refers to the average F1}
	\label{con:time_memory}
\end{table}
In Table~\ref{con:time_memory}, we compared the generation time on the whole testing set and used GPU memory on a single A100 across different LLMs. For the Phenotype Extraction task, we set the maximum input token length to 1500 and the maximum new token length to 50. In the Diagnosis Generation task, we set the maximum input token length to 1500 and the maximum new token length to 500. 
%%%%%%%%%%
We observed that although Mixtral 8*7B, LLaMA2 13B, and ClinicalCamel-70B exhibit better phenotype extraction performance, they have higher generation time and GPU memory usage. For example, ClinicalCamel-70B requires 2:50:16  for inference and uses 37,716 MB of GPU memory, while CancerLLM requires only 1:14:12 for inference and uses just 5,550 MB of GPU memory.

\subsection{Results of  Retrieval-based CancerLLM}

\begin{table*}[ht]
	\centering
	\renewcommand\arraystretch{1.3}
\resizebox{0.9\textwidth}{!}{%
\begin{tabular} {l|l|ccc|ccc|ccc|c}
\toprule 
    \multicolumn{1}{c}  {}&\multicolumn{1}{c}  {}&\multicolumn{3}{c}  {Exact Match} &\multicolumn{3}{c}  {BLEU-2}&\multicolumn{3}{c}  {ROUGE-L}&  F1\\
    %\hline
     Task&Retriever & Precision &  Recall & F1 & Precision &  Recall & F1 & Precision &  Recall & F1 & Average\\
      \midrule
      % \textbf{QA} \\
      \multirow{6}*{ Phenotype Extraction } & Random & 89.47 & 89.47 & 89.47 & 91.30 & 91.30 & 91.30 & 93.91 & 93.91 & 93.91&91.55\\
     & Medcpt & 87.95 & 87.95 & 87.95 & 90.18 & 90.18 & 90.18 & 92.97 & 92.97 & 92.97& 90.36\\
      &Contriever & 88.14 & 88.14 & 88.14 & 90.62 & 90.62 & 90.62 & \underline{94.04} & \underline{94.04} & \underline{94.04}&90.93\\
     & SGPT & 83.02 & 83.02 & 83.02 & 85.67 & 85.67 & 85.67 & 89.93 & 89.93 & 89.93&86.20\\
      &Specter2 & 88.61 & 88.61 & 88.61 & 90.81 & 90.81 & 90.81 & 93.67 & 93.67 & 93.67&91.03\\
          &No-retriever&  \underline{  89.37}    &   \underline{89.37 }      &   \underline{89.37}      &    \underline{91.98}     & \underline{91.98}    &   \underline{91.98}     &    93.98  & 93.98  &    93.98 &  \underline{91.78 } \\
      \midrule
      % \textbf{Diagnosis} \\
      \multirow{6}*{ Diagnosis Generation } & Random & 43.30 & 43.30 & 43.30 & 52.90 & 52.90 & 52.90 & 63.63 & 63.63 & 63.63&53.27\\
      &Medcpt & 58.34 & 58.34 & 58.34 & 65.99 & 65.99 & 65.99 & 73.54 & 73.54 & 73.54&65.95\\
      &Contriever & 82.50 & 82.50 & 82.50 & 85.15 & 85.15 & 85.15 & 88.80 & 88.80 & 88.80&85.48\\
     & SGPT & 43.94 & 43.94 & 43.94 & 52.97 & 52.97 & 52.97 & 63.75 & 63.75 & 63.75&53.55\\
      &Specter2 &\underline{ 85.78} &\underline{ 85.78} & \underline{85.78} & \underline{89.09 }& \underline{89.09} & \underline{89.09 }& \underline{92.49} & \underline{92.49} & \underline{92.49}&\underline{89.12}\\
           &No-retriever&    83.50     &   {83.50 }      &   {83.50}    & {86.60}      &{86.60}   &  {86.60}    &   {90.34}    &{90.34}   &{90.34}   &   {86.81}\\
   \bottomrule
\end{tabular}
}
\vspace{+2mm}
% ,  CancerLLM$^1$ refers to the model in the fine-tuning process using Biomixtrial 7B, while CancerLLM$^2$ refers to the model in the fine-tuning process using Mixtrial 7B. Bold font indicates the best model performance, while underline font represents our model's performance
\caption{Results of  retrieval-based CancerLLM  }
\label{con:Model_performance_rag}
\vspace{-1mm}
\end{table*}
As shown in Table~\ref{con:Model_performance_rag}. We evaluated the performance of CancerLLM with different retrievers on two tasks: phenotype extraction and diagnosis generation. For phenotype extraction, all retrievers achieved strong results, with Random and Specter2 performing slightly better than others, but the no-retriever baseline surprisingly achieved the highest average F1 score of 91.78\%. Contriever also showed competitive performance with an average F1 of 90.93\%, slightly outperforming MedCPT and SGPT. In contrast, diagnosis generation revealed a larger performance gap between retrievers. Specter2 led all models with an average F1 score of 89.12, followed by Contriever and the no-retriever baseline at 85.48\% and 86.81\%, respectively. Random and SGPT performed the worst in this task, with average F1 scores of only 53.27\% and 53.55\%. Overall, Specter2 and Contriever consistently delivered strong performance across both tasks, while the no-retriever baseline remained surprisingly competitive, especially in phenotype extraction.

%% file: 6_1_discussion.tex
\section{Discussion}
\label{con:discussion}

\subsection{Cancer Diagnosis Generation}
In this task, we primarily pre-train CancerLLM and explore its effectiveness in the diagnosis generation task by providing relevant information. As shown in Table~\ref{con:diagnosis_generation}, the Bio-Mistrial 7B obtains the best performance among all baseline LLMs. 
We guess that one reason for this is that Bio-Mistral 7B is trained on a large pre-processed corpus, such as the PubMed Central corpus~\cite{canese2013pubmed}. Another reason is that the foundational structure of Mixtral enhances training effectiveness. 
The Mixtral 8X7B outperforms the Mixtral 1X7B by more than 5\% in average F1 value. However, Mixtral 8X7B is a Mixture of Experts (MoE) model with 8 experts, making training and inference time challenging. Fortunately, our CancerLLM achieves the best model performance through pre-training and fine-tuning with a single expert model.
Not all LLMs with large parameters achieve the best performance. For example, the 70B models (LLaMA2 70B, Llama3-OpenBioLLM-70B, and ClinicalCamel-70B) do not outperform the Bio-Mistral 7B, despite being trained on medical corpora. This indicates that smaller parameter LLMs can achieve better performance.

% For ICDdiagnosis, we find that overall performance is lower than in diagnosis generation. This is because the expected diagnoses in ICD diagnosis are more complex.
% Our 7B model, CancerLLM, which incorporates cancer domain knowledge, also supports this point. This is particularly important in the medical domain as it reduces training time and memory usage, providing opportunities for medical institutions with limited computational resources.  

% \subsection{Cancer Treatment Plan Generation}
% By comparing extract match results of Table~\ref{con:Model_plan} and Table~\ref{con:diagnosis_generation}, we found that LLMs struggle with treatment plan generation, as even the best model, CancerLLM, achieves only 15.72\% performance. One reason is that the length of the ground truth treatment plan is longer than the diagnosis, making it difficult to generate the same treatment plan as created by doctors. Mistral 1*7B is also the most powerful model among all baselines, outperforming even the 70B models such as ClinicalCamel-70B. Mistral 1*7B achieves 52.42\% average F1 performance, whereas ClinicalCamel-70B achieves only 50.52\% average F1. It also provides evidence that smaller LLMs could achieve better performance compared to larger models. 
% % Despite the state-of-the-art model, CancerLLM, only achieved 55.40\%, it provides hope that LLMs can assist medical professionals in creating treatment plans.
% Increasing model parameters is not the sole method for enhancing model performance; exploring how to incorporate high-quality domain knowledge is also crucial.

\subsection{Cancer Phenotype Extraction}
In Table~\ref{con:Model_performance_phenotype},
We observed that the ClinicalCamel-70B model outperforms all baseline models in terms of average F1 value. However, its large parameter size significantly impacts both training and inference times, making it less efficient and more resource-intensive.
%%%%%%%%
On the other hand, our 7B model, CancerLLM, shows promising results, performing comparably to both the Mixture of Experts (MoE) model Mistrial 8*7B and the LLamA2 13B model. This is particularly noteworthy given CancerLLM's smaller parameter size, which makes it more efficient.
%%%%%%%%%%%%
We suspect that the reason behind CancerLLM's performance lies in the specific nature of the clinical notes and pathology reports we used for training. These documents may not contain sufficient annotation information, which is crucial for extracting phenotypic data. This highlights the importance of data quality and annotation in training effective LLM models for clinical information extraction tasks.

\subsection{Robustness}
The proposed two testbeds were primarily used to evaluate the counterfactual robustness and misspelling robustness of the LLMs. In the phenotype extraction task, when the counterfactual rate is set to 20\%, we observed that the performance of CancerLLM and Mixtrual 1*7B does not significantly decrease. However, when the rate exceeds 60\%, there is a notable decline in F1 performance. This indicates that Mixtrual 1*7B and CancerLLM can handle scenarios with fewer counterfactual instances. However, we found that when varying the misspelling rate, both the Bio-Mistral 7B and CancerLLM models experienced a significant decline in performance. This is particularly evident in the exact match metric, where performance drops close to zero. These results highlight the critical importance of correct spelling for LLMs to generate accurate diagnoses and treatment plans, as even minor misspellings can severely impact the models' ability to correctly interpret and process medical information. This underscores the necessity for meticulous data preprocessing and validation in clinical settings to ensure that the input data is free of errors, thereby enabling the models to function at their highest potential and provide reliable outcomes.

\subsection{Retrieval-based CancerLLM }
The results suggest that retrieval quality significantly impacts CancerLLM’s performance, especially in the diagnosis generation task. Specter2 consistently outperforms other retrievers, indicating its strength in retrieving clinically relevant and informative content. Interestingly, the no-retriever baseline performs competitively in phenotype extraction, suggesting that the model's inherent capabilities can handle simpler tasks without external context. However, for more complex reasoning tasks like diagnosis generation, external retrieval clearly provides substantial benefits. SGPT and Random retrievers underperform in both tasks, highlighting the importance of selecting domain-adapted retrieval methods. Overall, integrating high-quality retrievers like Specter2 or Contriever can significantly enhance CancerLLM’s effectiveness in clinical NLP applications.

\subsection{Error Analysis}

\subsubsection{Cancer Diagnosis Generation}
\begin{table}[ht]
	\centering
	\renewcommand\arraystretch{1.3}
	\scalebox{0.8}{
	\begin{tabular} {c|cc}
		\hline 
		Error type& Expected output& Error output\\ 
		\hline	
       \multirow{2}*{\makecell[l]{Incomplete Generation}}& \makecell[c]{\{breast and lung cancer\}\\ }  &  \makecell[c]{\{metastatic breast cancer\}\\ }  \\
       & \makecell[c]{\{nsclc t4n2mo-1a\}\\ }  &  \makecell[c]{\{nsclc t4n2mo\}\\ }  \\
       & \makecell[c]{\{metastatic lung cancer with brain mets\}\\ }  &  \makecell[c]{\{metastatic lung cancer\}\\ }  \\
         % \cmidrule(rr){1-3}
         \hline
          \multirow{1}*{\makecell[l]{Irrelevant Generation}}& \makecell[c]{\{ dcis (Ductal carcinoma in situ) \}\\ }  &  \makecell[c]{\{invasive ductal carcinoma\}\\ }  \\
         & \makecell[c]{\{nsclc\}\\ }  &  \makecell[c]{\{ lung cancer\}\\ }  \\
          
          \hline
           \multirow{1}*{\makecell[l]{Misspelling}}& \makecell[c]{\{ metastatic lung cnacer \}\\ }  &  \makecell[c]{\{metastatic lung cancer\}\\ }  \\
           \hline
            \multirow{1}*{\makecell[l]{Redundant Generation}}& \makecell[c]{\{ nsclc\}\\ }  &  \makecell[c]{\{nsclc,t2n0, s/p lobectomy and chemotherapy\}\\ }  \\
          & \makecell[c]{\{lung cancer\}\\ }  &  \makecell[c]{\{ lung cancer with brain mets\}\\ }  \\
           \hline
            \multirow{1}*{\makecell[l]{Abbreviation}}& \makecell[c]{\{ lung ca \}\\ }  &  \makecell[c]{\{lung cancer\}\\ }  \\
         
         \hline
	\end{tabular}
 }
		\caption{Error generation cases of diagnosis generation. Note: Due to privacy reasons, we did not provide the relevant clinical notes for each error type.}
	\label{con:error_cases_diagnosis_generation}
\end{table}
% As shown in Table~\ref{con:error_cases_diagnosis_generation}, we summarized five main error-generation cases of CancerLLM. We observed that: (1) Misspellings and abbreviations in the clinical notes affect the model's training and evaluation. (2) In some cases, the model tends to generate incomplete output, such as the ground diagnosis is the \textit{metastatic lung cancer with brain mets}, while the   model ignore the \textit{brain mets}.

As shown in Table~\ref{con:error_cases_diagnosis_generation}, we have comprehensively summarized two primary instances of error generation observed in CancerLLM. Our observations reveal the following:

\begin{itemize}
    \item (1) Misspellings and abbreviations within the clinical notes exert a significant influence on both the training and evaluation phases of the model. These linguistic inaccuracies can mislead the model's learning process and hinder its ability to accurately generate diagnoses.
   \item  (2) Another notable observation is the model's tendency to produce incomplete output in certain scenarios. For instance, when presented with a ground diagnosis such as \textit{metastatic lung cancer with brain metastases}, the model may overlook crucial details such as the presence of \textit{brain metastases}, This omission diminishes the comprehensiveness and accuracy of the generated diagnoses, potentially leading to misinterpretations and suboptimal clinical decisions.
\end{itemize}

%%%%%%%%%

These findings underscore the importance of meticulous data preprocessing and model refinement to mitigate the impact of linguistic nuances and enhance the CancerLLM's diagnostic capabilities.

\subsubsection{Cancer Phenotype Extraction}
\begin{table}[ht]
	\centering
	\renewcommand\arraystretch{1.3}
	\scalebox{0.8}{
	\begin{tabular} {c|lll}
		\hline 
		Error type& Input sentence  & Expected output& Error output\\ 
		\hline	
       \multirow{1}*{\makecell[l]{Redundant \\ Generation}}& \makecell[l]{the cores are infiltrated by ductal \\ carcinoma growing in nests and sheets \\with areas of residual lumen formation \\what is the histological type in the given context? }  & \makecell[l]{\{ductal carcinoma\}\\ }  &  \makecell[l]{\{ductal carcinoma, \\ 
      residual lumen\\ formation\}\\ }  \\
         \cmidrule(rr){1-4}
         \multirow{1}*{\makecell[l]{Irrelevant\\ Generation}}& \makecell[l]{one requisition slip and labeled right \\breast 1:00, part b on the container,\\ please describe the tumor location in the given context }  & \makecell[l]{\{1:00\}\\ }  &  \makecell[l]{\{it is not a  relevant \\question\}\\ }  \\
          \cmidrule(rr){1-4}
         \multirow{1}*{\makecell[l]{Repeat\\ Generation}}& \makecell[l]{In the text ypt2 ypn1bii mo, stage iib, \\what is the stage of cancer in the given context? }  & \makecell[l]{\{ypn1bii, iib,mo,ypt2\}\\ }  &  \makecell[l]{\{iib,iib,iib,iib,iib,\\iib,iib,iib\}\\ }  \\
           \cmidrule(rr){1-4}
        \multirow{2}*{\makecell[l]{Incomplete\\ Generation}}& \makecell[l]{In the text er/pr+, her2 negative, \\ please identify the receptors mentioned \\in the provided context}  & \makecell[l]{\{pr, her2, er\}\\ }  &  \makecell[l]{\{pr,er\}\\ }  \\
        & \makecell[l]{In the text sections from the stellate area in \\ the medical breast at the 9-10:00 position\\ show invasive, please describe the tumor location.}  & \makecell[l]{\{medical breast at the \\ 9-10:00\}\\ }  &  \makecell[l]{\{9-10:00 position\}\\ } 
        \\
   \cmidrule(rr){1-4}
        \multirow{1}*{\makecell[l]{Inaccurate \\ Annotation}}& \makecell[l]{In the text ductal carcinoma, solid, \\intermediate to high nuclear grade 3,\\ what is the grade of cancer?}  & \makecell[l]{\{high, intermediate\}\\ }  &  \makecell[l]{\{3\}\\ } 
        \\
         \hline
	\end{tabular}
 }
		\caption{Error generation cases of cancer phenotype extraction}
	\label{con:error_cases_phenotype_extraction}
\end{table}

% As shown in Table~\ref{con:error_cases_phenotype_extraction}, we summarized six main error-generation cases of CancerLLM. We observed that: (1) When contextual information is lacking, CancerLLM tends to generate repetitive and incomplete outputs. (2) Abbreviation is also a challenge for the model, such as in Table~\ref{con:error_cases_phenotype_extraction} (Missing generation), \textit{pr} and \textit{er}. (3) We find the CancerLLM is struggling to understand the question based on the given context, in Table~\ref{con:error_cases_phenotype_extraction} (Redundant generation), CancerLLM mistakenly identifies \textit{residual lumen formation} as the histological type. (4) Providing high-quality annotation data for phenotype extraction is crucial, as incorrect annotations can negatively impact the learning process of the LLM.

As shown in Table~\ref{con:error_cases_phenotype_extraction}, we have identified and summarized six primary error-generation cases, shedding light on crucial aspects influencing the model's performance. Our observations are as follows:
\begin{itemize}
    \item Redundant Generation: One prevalent issue is the model's tendency to generate repetitive and incomplete outputs when confronted with insufficient contextual information. This limitation hampers the model's ability to provide comprehensive and accurate responses.

    \item Abbreviation Challenges: Abbreviations pose a significant challenge for CancerLLM, as evidenced by instances such as those outlined in Table~\ref{con:error_cases_phenotype_extraction} (Incomplete generation), where abbreviations like \textit{pr} and \textit{er} lead to incomplete outputs. Resolving this challenge is vital for enhancing the model's interpretive accuracy.

\item Contextual Misinterpretation: CancerLLM struggles to comprehend questions accurately within the given context, as illustrated in Table~\ref{con:error_cases_phenotype_extraction} (Redundant generation), where it misidentifies \textit{residual lumen formation} as a histological type. This highlights the need for improved contextual understanding to prevent such errors.

\item Annotation Quality: The quality of annotation data for phenotype extraction significantly impacts CancerLLM's learning process. Inaccurate annotations, as discussed, can impede the model's ability to learn effectively, underscoring the importance of providing high-quality annotation data to facilitate robust learning outcomes.

\end{itemize}

These findings emphasize the multifaceted nature of the challenges faced by CancerLLM and underscore the importance of addressing these issues to enhance its performance and reliability in clinical applications.

%% file: 5_method.tex
\section{CANCERLLM}
\begin{figure*}[t]
        \centering
        \includegraphics[width=1\columnwidth]{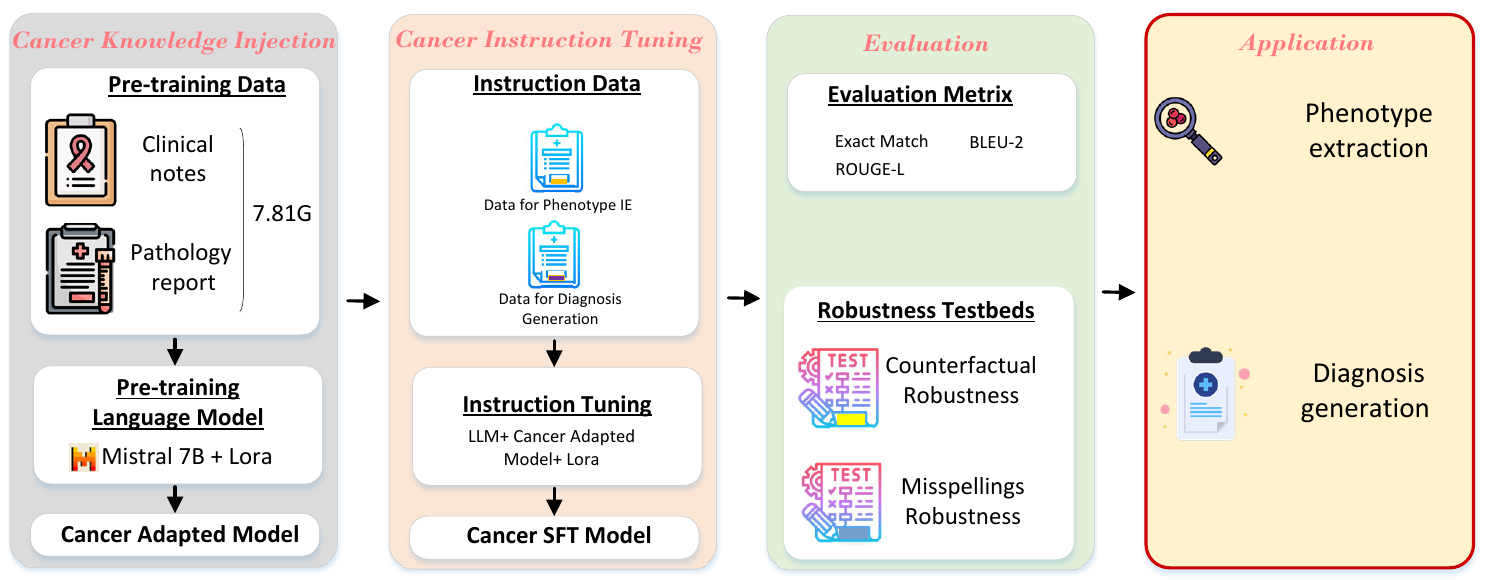}
	\caption{ Overview of \textsc{CancerLLM}} 
	\label{con:whole_framework}
\end{figure*}
Figure~\ref{con:whole_framework} illustrates the workflow of the \textsc{CancerLLM}, beginning with the injection of cancer knowledge and followed by supervised instruction tuning. Subsequently, we assess the generative capability of our framework compared to current medical LLMs using three evaluation metrics. Additionally, we introduce two testbeds to evaluate the robustness of the LLMs. Finally, the trained \textsc{CancerLLM} is applied to two specific tasks.

\subsection{Pre-training Data for Cancer}
The data used in this study were obtained from the University of Minnesota (UMN) Clinical Data Repository. It contains the health records of 31,465  cancer patients. Specifically,  It includes 2,676,642 cancer clinical notes (7.27GB) and 515,524 cancer pathology reports (536.85MB).
We obtained the data with the approval of the UMN Institutional Review Board (IRB).

% The statistics information of patient and clinical notes are shown in Figure~\ref{con:Notes_Statistics}.
% under 
% STUDY00017350: Computational Models for Cancer.
% \#1210M22601.

% \begin{figure}[htbp]
%     \centering
%     % \subfigure[Number of patients with different cancers]{
%     %     \includegraphics[width=1.8in]{cancer_LLM_patient_1.pdf}
%     % }
%      \subfigure[Number of patients with different cancers]{
%         \includegraphics[width=4in]{cancer_LLM_patient_2.pdf}
%     }
%     %  \subfigure[Number of clinical notes with different cancers]{
%     %     \includegraphics[width=2in]{cancer_LLM_patient_notes_1.pdf}
%     % }
%      \subfigure[Number of clinical notes with different cancers]{
%         \includegraphics[width=4in]{cancer_LLM_patient_notes_2.pdf}
%     }
    
%     % \quad    %用 \quad 来换行
%     \caption{Patients and clinical notes Statistics}
%    \label{con:Notes_Statistics}
% \end{figure}

\subsection{Instruction Tuning Data for  Cancer}
To adapt \textsc{CancerLLM} for various downstream applications, we employ instruction tuning for two distinct tasks: cancer phenotypes extraction, cancer diagnosis generation. This section provides detailed information on each task description and its instruction tuning data.

\subsubsection{Cancer Phenotypes extraction}
\textbf{Task description} By providing sentences from cancer pathology reports or cancer clinical notes, the model needs to extract eight specific entities: hormone receptor type, hormone receptor status, tumor size, tumor site, cancer grade, histological type, tumor laterality, and cancer stage.

\textbf{Dataset}
The dataset (CancerNER) used in CancerBERT~\citep{zhou2022cancerbert} focuses on named entity recognition (NER) of breast cancer-related phenotypes. This data is extracted from both clinical notes and pathology reports within electronic health records (EHRs). By inputting a sentence, CancerBERT is expected to identify and recognize each entity corresponding to the aforementioned types.  LLMs have been found to underperform in NER tasks compared to supervised models tailored for sequence labeling \citep{wang2023gpt}. This is primarily due to the inherent differences between sequence labeling tasks, which require the model to tag each token in a sequence, and the generation tasks that LLMs are typically optimized for generating the next tokens.
So, in this paper, to enhance the performance of NER, we transformed the CancerNER dataset into a question-answering format by posing eight questions based on the entity types, the answers are relevant entities. The question types are shown in Table~\ref{con:qa_dataset}. Note: by giving the input sentence (context), when the question is not relevant with the context, our model is expected to output "Not relevant". 
% Finally, this dataset includes 4,614 training instances, 1,054 testing instances, and 980 development instances
. Each instance follows the template \{instruction, context, response\}. The instruction is: \textit{You are a medical expert, this task involves answering the question based on the provided context or text.}, The context is the sentence that needs to be processed, along with a question. The response is the identified entity.
\begin{table}[ht]
	\centering
	\renewcommand\arraystretch{1.3}
	\scalebox{0.9}{
	\begin{tabular} {cc}
		\hline 
		& question \\ 
		\hline	
        1 &  What is the tumor size in the given context?  \\
    2&  What is the histological type in the given context? \\
     3& Please identify the receptors mentioned in the provided context. \\
      4&  What is the receptor type in the given context? \\
       5&  Please identify the value of tumor laterality in the provided context. \\
        6&  What is the stage of cancer in the given context? \\
        7& Please describe the tumor location in the given context  \\
        8& What is the grade of cancer in the given context?\\
		\hline
	\end{tabular}
 }
		\caption{Question types}
	\label{con:qa_dataset}
\end{table}

\subsubsection{Diagnosis Generation}
\textbf{Task description} By giving the information from cancer clinical notes,  which includes the 1) reason for visit, 2) treatment site, 3) subjective information, 4) nursing Review of Systems (ROS), 5) objective observations, and 6) laboratory test results, the model is expected to generate the correct cancer diagnosis.

% Our testing dataset includes 370 different diagnosis results,  such as \textit{p(anal) cancer, breast cancer, malignant neoplasm, cancer of the vocal cords, and txn2amo(iiia) triple negative.}

\textbf{Dataset}
We randomly selected 10,635 cancer clinical notes that do not overlap with the pre-training dataset, and split them into 80\% instances for training, 10\% instances for testing, and  10\%  instances for validation. These datasets do not overlap.
Each instance follows the template \{instruction, context, response\}. The instruction is: \textit{You are a medical expert. This task involves generating the diagnosis based on the provided context or text}, the context includes the reason for visit, treatment site, subjective information, nursing Review of Systems (ROS), objective observations, and laboratory test results. The response is the cancer diagnosis. In general, when recording a diagnosis in the clinical notes, doctors may describe the same diagnosis using different terms or abbreviations, such as 'L lung cancer' to refer to 'left lung cancer.' To standardize these diagnoses, ICD codes are introduced. However, mapping these varied descriptions to a unified ICD code is a challenging task. To evaluate our model's performance, we also created a test dataset containing 374 clinical notes, where the ground truth diagnosis is represented by the ICD code stored in the database. This testing dataset is called ICDdiagnosis.

\subsection{Training Methodology}
Our training methods include two phases: continued pre-training and instruction tuning. In this work, we choose Mistral 7B~\citep{jiang2023mistral} as the foundation model to inject cancer knowledge in the pre-training progress, as it has shown superior performance on various medical benchmarks. To improve the efficiency of the training process, we utilize Low-Rank Adaptation (LoRA)~\citep{hu2021lora}.
\subsubsection{Continued Pre-training}
This stage employs Mistral 7B, focusing on next-token prediction on clinical notes and pathology reports. 
Due to computing resource limitations, we undergo pre-training using LoRA.
% The total number of parameters is 7,262,703,616. By using LoRA, 28.87\% of the parameters are trainable (20,971,520 parameters).
We parallelize the computation across ten 80G A100 GPUs, with each A100 utilizing around 40G of GPU memory. The total training time is 47 hours. The training batch size and evaluation batch size per device are set to 4. For LoRA, the rank is set to 8, alpha is set to 16, and dropout is set to 0.05 to optimize learning. We use the AdamW optimizer and set the learning rate to 2e-4. The training epoch is 1, and the evaluation step is 0.2.
\subsubsection{Instruction Tuning}
After continued pre-training, our model undergoes instruction tuning with two datasets, aligning with different clinical requirements. We also use LoRA to fine-tune the model. The rank is set to 64, and alpha is set to 16. The training step is 5. We use the AdamW optimizer~\citep{loshchilov2017decoupled} with a learning rate of 2e-4. The model was fine-tuned using a single A100 GPU with 4 hours of training time and 7338M of GPU memory.
\subsection{Baselines and Evaluation Metrics}
In our work, we used 16 widely used clinical, biomedical, and general LLMs as the baseline models, including: seven 7B LLMs ( PMC LLaMA 7B~\citep{wu2023pmc}, Medalpaca 7B~\citep{han2023medalpaca}, LLAMA-2 7B~\citep{han2023medalpaca}, Mistral 1*7B~\citep{jiang2023mistral}, Mixtral 8*7B~\citep{jiang2023mistral} , Qwen-7B~\citep{bai2023qwen}, and Bio-Mistral 7B~\citep{labrak2024biomistral} ), two 8B LLM ( LLama3 8B\footnote{https://huggingface.co/meta-llama/Meta-Llama-3-8B}, Deepseek 8B\footnote{https://huggingface.co/deepseek-ai/DeepSeek-R1-Distill-Llama-8B}), 
four 13B models (MedLLaMA 13B~\cite{wu2023pmc} , PMC LLaMA 13B~\citep{wu2023pmc},  Medalpaca 13B~\citep{han2023medalpaca}, and   LLaMA2 13B~\citep{touvron2023llama} ) three 70B LLMs ( LLaMA2 70B~\citep{touvron2023llama} , Llama3-OpenBioLLM-70B\footnote{https://huggingface.co/aaditya/Llama3-OpenBioLLM-70B} and  ClinicalCamel-70B~\citep{toma2023clinical} ).
Same with ~\cite{li2023understand},  we evaluate all the models based on generative evaluation metrics, including
Exact Match~\citep{rajpurkar2016squad} , BLEU-2~\citep{papineni2002bleu}, and ROUGE-L~\citep{cohan2016revisiting}.

\subsection{Robustness Testbeds}

\paragraph{Counterfactual Robustness}
Constructing a high-quality annotation corpus is challenging work for phenotype extraction, as it often involves dealing with incorrect data labeling. 
In our work, the mislabeled instances are called counterfactual instances.
%%%%%%%
In the condition of the mislabeled training dataset, the LLM may have the ability to avoid negative information.
%%%%%%%%%%%%%
To validate the counterfactual robustness, same as ~\cite{li2024benchmarking}, we introduced the counterfactual robustness testbed. Specifically, when constructing the training dataset of phenotypes extraction, we set the negative rate to be $20\%$, $40\%$, $60\%$ and $80\%$, corresponding to $20\%$, $40\%$, $60\%$ and $80\%$  of instances being wrongly labeled. An example of incorrect annotation in this dataset would be entities or entity types that are not present in the input sentence or are irrelevant to the entity.
Subsequently, the input sentence along with the instruction is fed into the LLM to generate the output.

\paragraph{Misspellings Robustness}
By checking the clinical notes, we found there are some misspelling words, such as \textit{diagnosis} is written to \textit{dinosis}. To validate the Misspellings robustness of LLM,  we introduced the misspellings robustness testbed. Specifically, when constructing the training dataset for diagnosis generation and treatment plan generation, we set the misspelling rate to be $2\%$, $4\%$, $6\%$, $8\%$, corresponding to $2\%$, $4\%$, $6\%$, $8\%$ of the words in a sentence being misspelled.  
Subsequently, the input sentence along with the instruction is fed into the LLM to generate the output.

\section{Retrival-based CanerLLM}

To further improve performance on cancer-related tasks and evaluate the influence of different examples in instruction tuning, we propose Retrieval-based CancerLLM. In this model, five retrievers—Random, MedCPT~\citep{jin2023medcpt} , Contriever~\citep{izacard2021unsupervised}, SGPT~\citep{muennighoff2022sgpt}, and Specter2~\citep{Singh2022SciRepEvalAM} are employed to retrieve relevant documents for each input sentence from the corpus. Following the approach in \cite{li2024rt}, we use the training set as the retrieval corpus. More specifically, the Retrieval-based CancerLLM contains three main steps,

\begin{itemize}
    \item Step 1) Vector Representation of Input Sentence: The retrievers first compute the vector representation of the given input sentence using their respective retrieval models. These retrievers rely on dense embeddings generated through neural networks.

    \item Step 2)  Vector Computation for Corpus Instances: Each retriever then calculates the vector representation for every instance in the retrieval corpus. They also employ deep learning-based embeddings to encode semantic information.

    \item Step 3)  Similarity Computation and Retrieval: Finally, the retrievers measure the similarity between the input sentence vector and each instance vector in the corpus. The most relevant instances, determined by the highest similarity scores, are selected and fed into the CancerLLM alongside the input sentence as examples, helping the model generate more accurate predictions.
    
\end{itemize}

%% file: 7_conclution.tex
\section{Conclusion}

In this study, we introduced CancerLLM, a medical LLM designed specifically for the cancer domain. We provided two datasets for cancer phenotype extraction, cancer diagnosis generation. Additionally, we proposed two testbeds to evaluate the robustness of CancerLLM. Comparing CancerLLM with 16 widely used LLMs, our results demonstrate that CancerLLM achieves state-of-the-art performance across three tasks. To enhance model performance, we also investigate the effectiveness of different retrievers. Our work provides invaluable insights and tools for further research on leveraging AI to enhance cancer domain.

% \section{Limitations}

% In this study, we utilized the training set as the retriever corpus for the question-answering task. However, several studies utilize larger corpora with richer knowledge, such as PubMed and Wikidata. Consequently, in other tasks such as link prediction, augmenting the size of the labeled corpus remains a formidable challenge. Additionally, three retrievers select the most relevant instance of the input sentence as an example, taking into account factors like the LLM input window size, training efficiency, and inference efficiency. Including too many examples beyond the window size could potentially impact inference efficiency. Overcoming the limitations imposed by window size and ensuring efficient inference remains a challenge.

\section{Acknowledgments}
This research was supported by the National Institutes of Health’s National Center for Complementary and Integrative Health grant number R01AT009457, National Institute on Aging grant number R01AG078154 and National Cancer Institute grant number R01CA287413. The content is solely the responsibility of the authors and does not represent the official views of the National Institutes of Health.   We thank support from UMN's Center for Leanring Health System Sciences. We thank support from UMN Data Science Initiative seed grant.
%%%%%%%%
This research was supported by the National Institutes of Health’s National Center for Advancing Translational Sciences, grant UM1TR004405. The content is solely the responsibility of the authors and does not necessarily represent the official views of the National Institutes of Health’s National Center for Advancing Translational Sciences.

% Thanks to Chad Dupuis for solving the issues with our GPU server.  

% \section{Author Contributions}

% \begin{itemize}
%   \item Mingchen Li:  Digging for ideas, doing related work, experiments, coding,  Methodology, Model Design, Software, Writing – original draft, Paper revision.
%   \item Halil Kilicoglu:  Paper revision.
%   \item Hua Xu:  Paper revision.
%    \item Rui Zhang: Supervision, idea and method exploration, Writing – original draft, Writing – review \& editing. 
% \end{itemize}

\section{Competing Interests}
The authors declare no competing financial or non-financial interests.